\documentclass[conference]{IEEEtran}
\IEEEoverridecommandlockouts
\usepackage{cite}
\usepackage{amssymb,makecell}
\usepackage{graphicx}
\usepackage[normalem]{ulem}
\usepackage{textcomp}
\usepackage{xcolor}
\usepackage{amsmath,amsfonts}
\usepackage{algorithmic}
\usepackage{algorithm}
\usepackage[algo2e]{algorithm2e} 
\usepackage{array}
\usepackage{textcomp}
\usepackage{stfloats}
\usepackage{multirow}
\usepackage{url}
\usepackage{verbatim}
\usepackage{array}
\usepackage{multirow}
\usepackage{longtable}
\usepackage{rotating}
\usepackage[caption=false]{subfig} 
\newcommand{\etal}{\textit{et al}. }

\def\BibTeX{{\rm B\kern-.05em{\sc i\kern-.025em b}\kern-.08em
    T\kern-.1667em\lower.7ex\hbox{E}\kern-.125emX}}
\begin{document}

\title{On the Design of Communication-Efficient Federated Learning for Health Monitoring\\
\thanks{D. Chu is with the School of Information and Communication Engineering of the University of Electronic Science and Technology of China, Chengdu, China, email: dong.chu.uestc@gmail.com. W. Jaafar is with the Software and Information Technology Engineering department of \'{E}cole de Technologie Sup\'{e}rieure, QC, Canada, email: wael.jaafar@etsmtl.ca. H. Yanikomeroglu is with the Systems and Computer Engineering department of Carleton University, ON, Canada, email: halim@sce.carleton.ca.}
}

\author{\IEEEauthorblockN{Dong Chu, Wael Jaafar, and Halim Yanikomeroglu}
}

\maketitle

\begin{abstract}
With the booming deployment of Internet of Things, health monitoring applications have gradually prospered. Within the recent COVID-19 pandemic situation, interest in permanent remote health monitoring solutions has raised, targeting to reduce contact and preserve the limited medical resources. Among the technological methods to realize efficient remote health monitoring, federated learning (FL) has drawn particular attention due to its robustness in preserving data privacy.
However, FL can yield to high communication costs, due to frequent transmissions between the FL server and clients. 
To tackle this problem, we propose in this paper a communication-efficient federated learning (CEFL) framework that involves clients clustering and transfer learning. First, we propose to group clients through the calculation of similarity factors, based on the neural networks characteristics. Then, a representative client in each cluster is selected to be the leader of the cluster. Differently from the conventional FL, our method performs FL training only among the cluster leaders. Subsequently, transfer learning is adopted by the leader to update its cluster members with the trained FL model. Finally, each member fine-tunes the received model with its own data. To further reduce the communication costs, we opt for a partial-layer FL aggregation approach. This method suggests partially updating the neural network model rather than fully. Through experiments, we show that CEFL can save up to to 98.45\% in communication costs while conceding less than 3\% in accuracy loss, when compared to the conventional FL. Finally, CEFL demonstrates a high accuracy for clients with small or unbalanced datasets.
\end{abstract}


\vspace{-10pt}
\section{Introduction}
Internet of Things (IoT) technology has raised in recent years allowing its application in several areas such as e-health wearable devices, smart homes, autonomous cars, etc. IoT technology has been constantly improving our lives, and one of the most rapidly developing IoT services is health monitoring. Indeed, IoT sensors can be used to observe a patient's condition, detect early an illness, or alert the medical staff about a critical health condition \cite{selvaraj2020challenges}. 
Within a pandemic situation where the medical staff is constantly under pressure, remote health monitoring has been rapidly developing to partially alleviate this burden. Supported by IoT devices, remote health systems can reliably ensure self-treatment at home, detect and monitor emergency situations such as a heart attack or falling of an elderly, and automatically calling for assistance from the adequate first responder staff \cite{rahman2020defending}.

In order for these services to be efficient, data need to be collected from IoT devices, filtered, and processed. For instance, to predict a falling event, motion data need to be analyzed, while electroencephalography (EEG) signal and heart rate history data can serve to synthesize an in-depth report and pre-diagnose an illness.
Data processing would typically rely on cloud computing platforms \cite{verma2018cloud}. To make remote health monitoring more accurate, large-scale machine learning (ML) approaches can be leveraged. However, the associated centralized data and model training process bring serious data security and privacy concerns.


Rather than learning from centrally collected user data and being exposed to the risk of privacy leakage, federated learning (FL) can address this problem using a collaborative model through the communication of only the training model, while keeping the training process and data at the local level, i.e., close to patients. 
However, FL raises other concerns, such as high communication costs and systems heterogeneity
\cite{li2020federated}. Indeed, frequent communications can rapidly become the bottleneck of the FL development. This is caused mainly by the important number of communication rounds between the server and clients and the size of transmitted data. 

In this context, we aim here to reduce the communication costs of FL, applied for a specific health monitoring service example, i.e., patients activity detection. To do so, we propose the integration of graph clustering and transfer learning techniques into FL, which would drastically reduce the data exchange rounds. Specifically, FL is realized among only a fraction of the available clients with highly significant data features, and with limited data exchange. To the best of our knowledge, this work is among the first ones that combines such techniques in order to reduce the communication costs of FL. Subsequently, the main contributions of this paper can be summarized as follows:
\begin{itemize}
	\item We propose to cluster FL clients based on their mutual similarity, measured from their neural network weights, then we select a cluster leader for each one of them.
    \item Next, we propose to perform federated learning among only a fraction of the clients, i.e., cluster leaders, thus cutting down the number of clients participating in FL.
    \item To further reduce the communication costs,
    we opt for partial-layer FL aggregation, where we select the weights representing the most interesting FL features only.

    \item Through experiments, we demonstrate the efficiency of our FL method in drastically reducing the communication costs at the expense of a slight loss in FL accuracy, compared to the conventional FL approach. 
\end{itemize}

The rest of the paper is organized as follows. Section \ref{sec 2} reviews related works. Section III describes the conventional FL system. Section \ref{sec 3} presents the proposed FL framework. In Section \ref{sec 4}, experimental results are provided to evaluate the performances of our FL approach and validate its efficiency from both the accuracy and communication cost perspectives. Finally, Section \ref{sec 5} concludes the paper.

\vspace{-5pt}
\section{Related Works}
\label{sec 2}

There is a growing need for health monitoring to be cost-efficient, reliable, and accessible. Thus, federated learning, as a subdivision of machine learning that guarantees data privacy, is suitable for healthcare applications. 

For electronic health records, Liu \etal proposed federated-autonomous deep learning to train different parts of the FL model using all or specific data sources \cite{liu2018fadl}. 
To cope with the unbalanced distributed datasets in the edge computing system, Duan \etal built a self-balancing FL framework that uses data augmentation and multi-client rescheduling \cite{duan2019astraea}. 
Similarly, a cloud-edge based FedHome framework was proposed by Wu \etal to handle the unbalanced and non-independent and identically distributed data via the generative convolutional autoencoder, thus realizing accurate and personalized health monitoring \cite{Wu2020}.
Moreover, an efficient activity recognition application based on FL has been developed in \cite{sozinov2018human}. FL was used to mitigate the privacy violation problem and to reduce data collection costs for centralized training, while 
Fang \etal proposed in \cite{fang2020highly} privacy preservation and communication costs reduction through the use of a lightweight encryption protocol. 
Focusing on the FL communication efficiency only, researchers proposed compression-based methods to reduce the size of the communicated model. For instance, Konečný \etal proposed to reduce the size of the uplink data  through structured and sketched updates, where an update is learned from a restricted parameterized space and compressed prior to upload \cite{konevcny2016federated}. Also, sparse ternary compression was proposed in \cite{sattler2019robust}, which is proved to be more robust and converge faster than the federated averaging benchmark. Other communication cost reduction approaches include FedPAQ \cite{Amirho2020} that allows only partial device participation and periodic server averaging for quantized message uploads, and CMFL \cite{luping2019cmfl}, which reduces the number of updates by eliminating irrelevant ones to the global model improvement tendency.


%
%
%

\section{FL Preliminaries}
In the conventional federated learning, users train a global neural network model collaboratively without having to share their local data. FL aims to realize an empirical global optimization through the iterative global aggregation and edge model update. For a system with $N$ clients, let $D_n$ be the dataset of client $n$ and $f_i(w)$ the loss minimization objective of sample $i$. The objective is to minimize the training loss function $F_n(w)$ for client $n$, where
\begin{equation}
\small
    F_n(w) = \frac{1}{|D_n|}\sum_{i\in D_n} f_i(w),
    \label{eq9}
\end{equation}
where $|D_n|$ is the number of data samples in $D_n$. In each FL training round $t$, participating clients get from the FL server the latest global neural network model $\omega(t)$. Then, each client executes a number of local training episodes $\varepsilon$ based on its local data. At the end of $\varepsilon$ episodes, each client sends its local model $\omega({t+1})$ to the server, and the latter aggregates all received local models into its own global model as follows:
\begin{equation}
    \small
    F(w)=\sum_{n=1}^N \frac{|D_n|}{|D|} F_n(w),
    \label{eq10}
\end{equation}
where $|D|$ denotes the number of data samples from all clients.
This process describes one global FL round, where the conventional optimization objective of FL is given by (\ref{eq10}).


Unlike conventional FL, we present next our proposed method for communication costs reduction in FL, called communication-efficient FL (CEFL).

\section{Proposed CEFL Framework}
\label{sec 3}


\begin{figure}[t]
	\centering
	\includegraphics[width=0.48\textwidth]{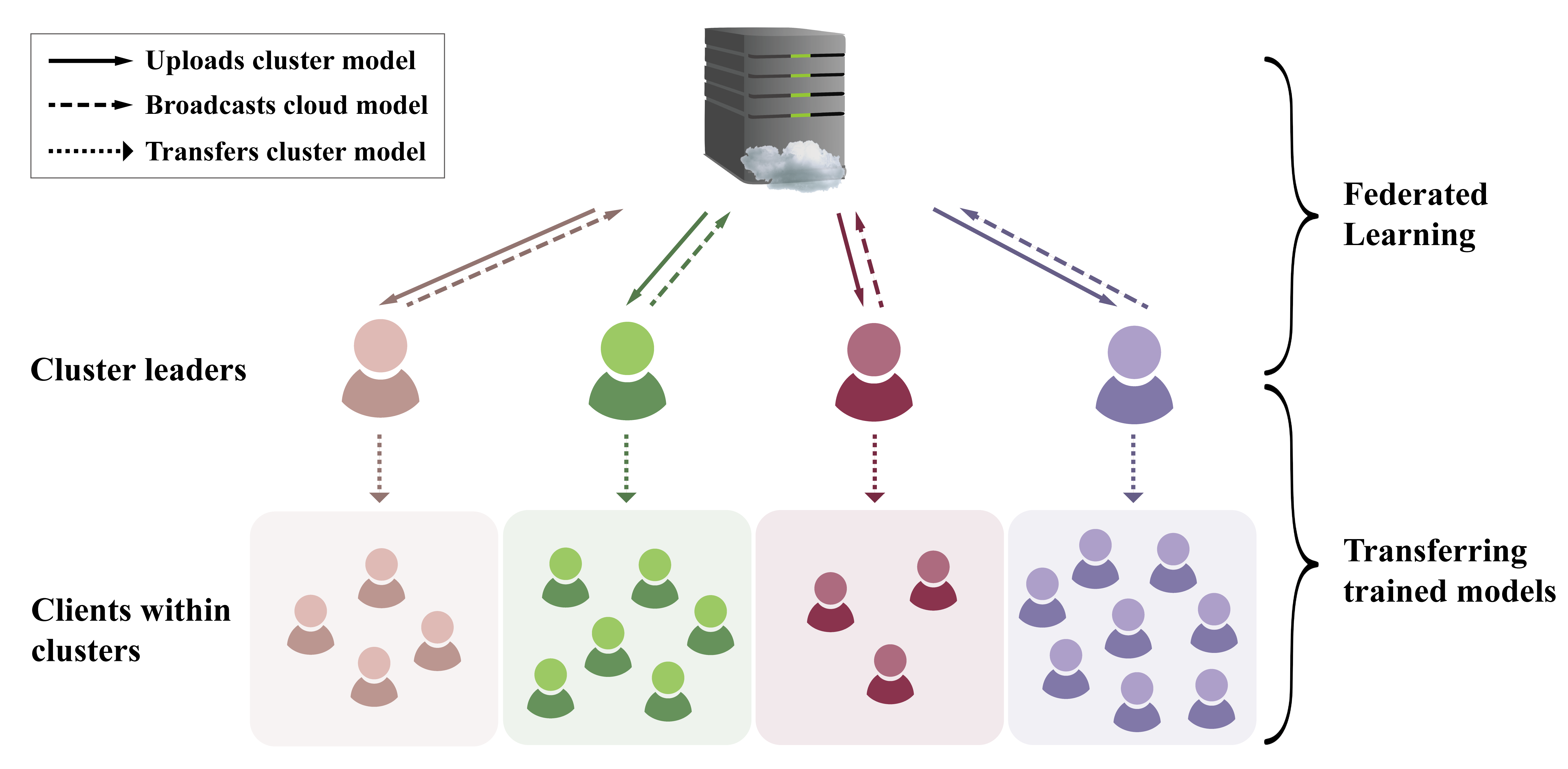}
	\caption{The CEFL framework. 
	}
	\label{fig1}    
\end{figure}

The CEFL is depicted in Fig. \ref{fig1} where we distinguish between two training sessions, namely FL and transfer learning, which are detailed below.

\vspace{-5pt}
\subsection{Federated Learning Session}
\begin{figure*} 
	\centering
	\subfloat[Similarity graph]{
		\includegraphics[width=0.35\linewidth]{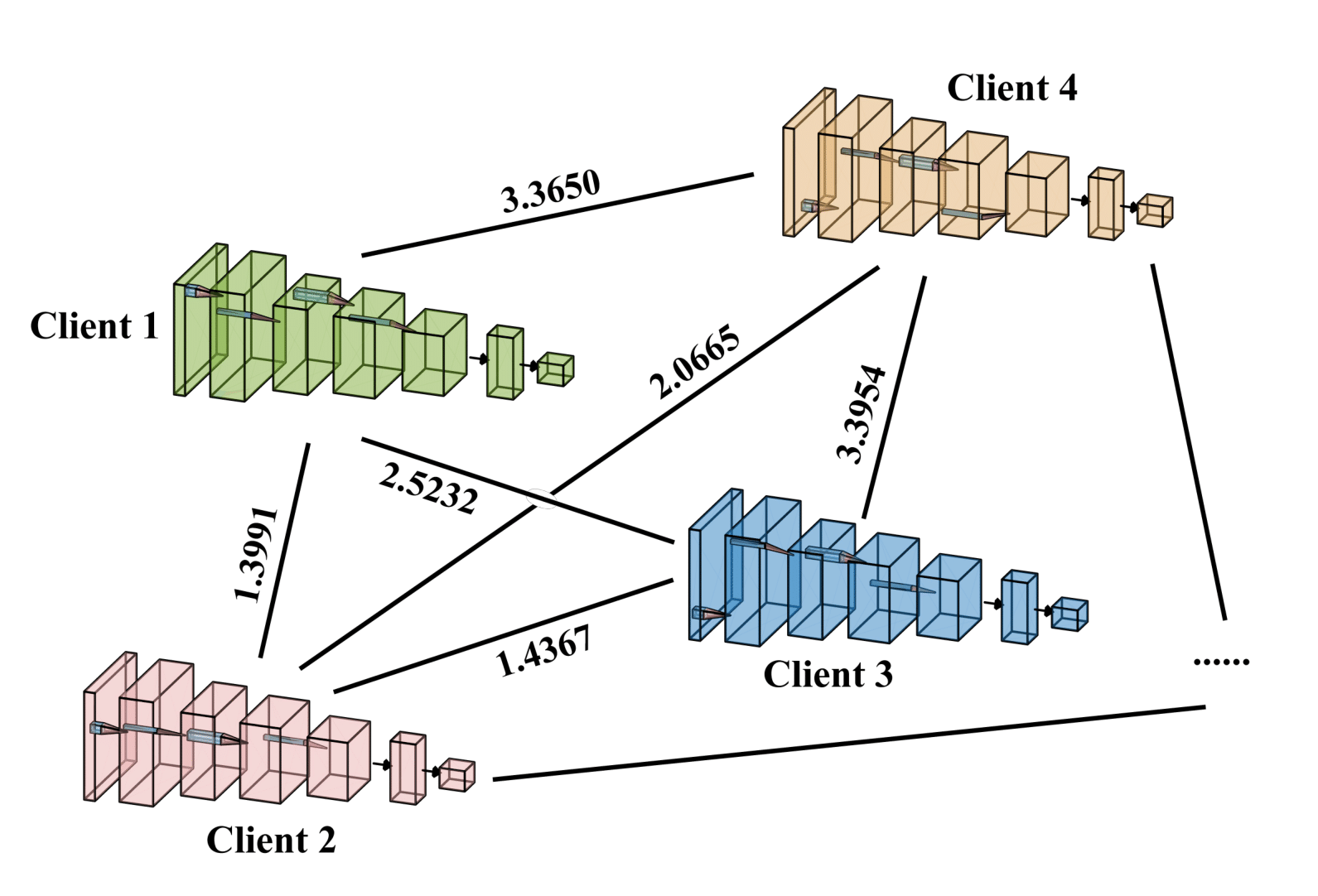}}
	\label{2a}\hfill
	\subfloat[Similarity graph prior to clustering]{
		\includegraphics[width=0.21\linewidth]{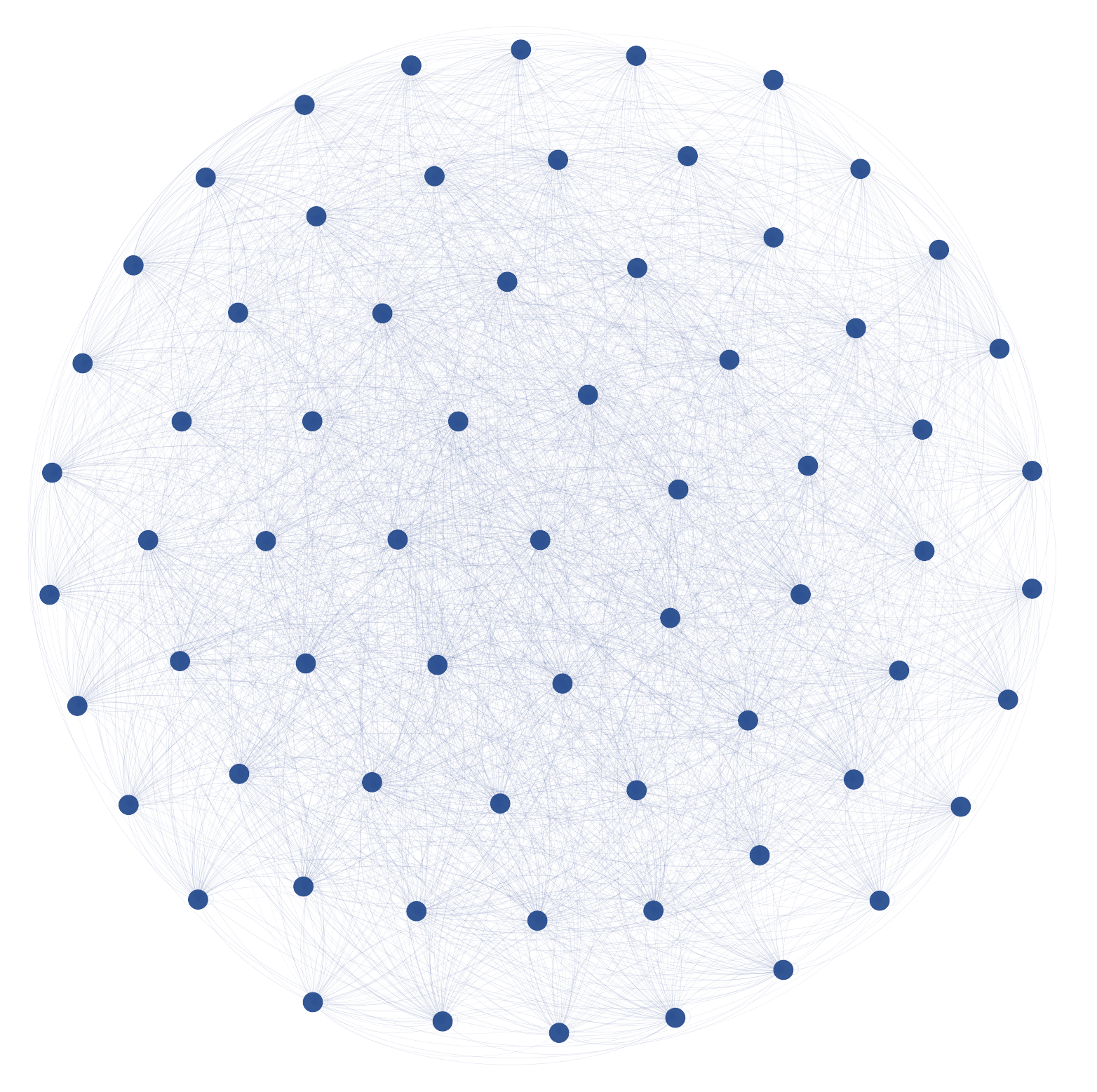}}
	\label{2b}\hfill
	\subfloat[Clustering result]{
		\includegraphics[width=0.21\linewidth]{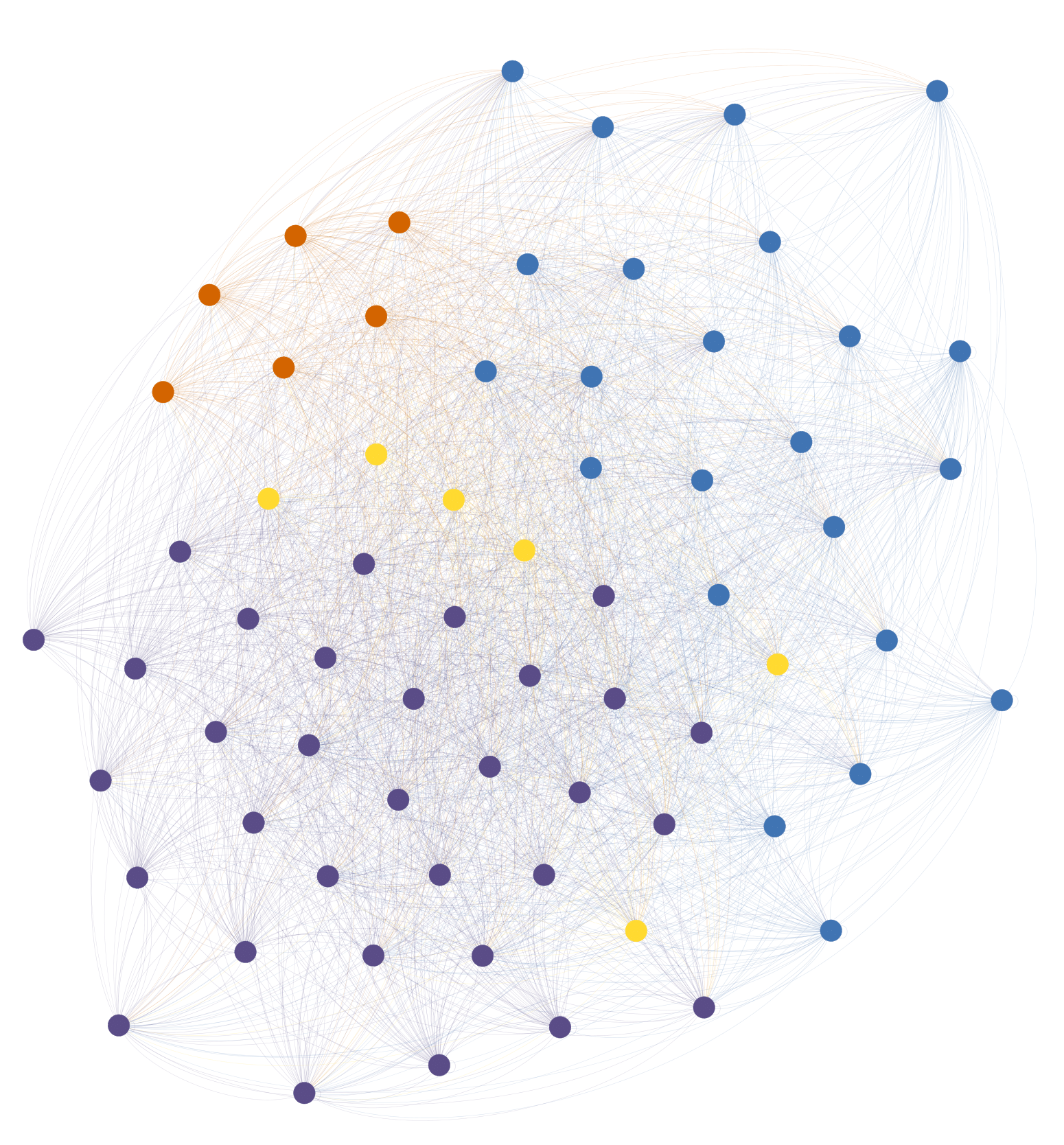}}
	\label{2c}\hfill
	\caption{Clustering clients based on similarity: (a) Building the similarity graph among clients. Each client is represented with its neural network model. The edges' values are the similarity factors calculated with eq.(\ref{eq2}). (b) A different representation of the similarity graph prior to clustering. Each node is a client. (c) Clustering outcome. Nodes with different colors represent clients clustered together.}
	\label{fig2} 
\end{figure*}

Before running FL rounds, four steps need to be executed: 

\begin{itemize}
	\item \textbf{Step 1 (Building the clients' similarity graph):} We begin by quantifying the clients' mutual similarity. First, we train the local models for a small number of episodes and extract their neural network weights. Then, to evaluate the similarity factor of two clients $i$ and $j$, denoted $d_{ij}$, we calculate the Euclidean distances between their weights corresponding to the same network layer, and sum them over all layers:
	\begin{equation}
\small
		d_{ij} = \sum_{l=1}^{L} \left\|\omega_i^l - \omega_j^l\right\|,
	\end{equation}
	where $L$ is the number of neural network layers for the clients' model, $\omega_i^l$ is the set of neural network weights at the $l^{th}$ layer of client $i$'s model, and $||\cdot||$ is the Euclidean distance operator.
	Consequently a graph $G(V,E)$ can be built, where vertices $V$ and edges $E$ represent the clients and similarity factors, respectively. For accurate representation within the graph, we assign the weights $S_{ij}$ to the edges rather than $d_{ij}$, where
	\begin{equation}
	\small
		S_{ij} = -d_{ij} + d_{\min} + d_{\max},
		\label{eq2}
	\end{equation}
	and $d_{\min}$ and $d_{\max}$ are the minimum and maximum values of $d_{ij}$, $\forall i,\forall j$, respectively. Hence, a large $S_{ij}$ refers to high similarity and small $S_{ij}$ to low similarity. In Fig. \ref{fig2}.a, a similarly graph example is illustrated.
	
	
	
	\item \textbf{Step 2 (Clients clustering):} Given the similarity graph, we adopt the Louvain algorithm to detect community structures, i.e., clients with strong similarities, within $G(V,E)$ \cite{blondel2008fast}. Our choice of clustering algorithm is motivated by its fast convergence, implementation simplicity, and customizability. 
    The Louvain algorithm is a greedy approach that allows to optimize the modularity as it runs. The modularity score (between -0.5 and 1) measures the relative density of edges inside communities with respect to those outside communities.  When using the Louvain algorithm, the number of clusters needs to be specified according to the demand for cluster sizes.
	In Figs. \ref{fig2}.b and \ref{fig2}.c, we depict the graph clustering step.
	
	\item \textbf{Step 3 (Leader selection):} Following step 2, we designate one client to be the leader of the cluster. Its responsibility consists on participating in the FL session.
	The leader is selected as the one sharing the highest similarity with the clients of its cluster. In other words, a client $c_k$ is the leader of cluster $k$ only if
	\begin{equation}
	\small
		c_k = \arg \max_{i} \sum_{j\in C_k; j\neq i}S_{ij}, 
		\label{eq3}
	\end{equation}
	where $C_k$ is the set of clients in cluster $k$. The above steps lay the foundation for FL among the cluster leaders. 
    \item \textbf{Step 4 (Partial-layer FL aggregation):} Instead of the conventional FL that updates the whole neural network model weights, we opt here for a partial aggregation strategy, aiming to preserve more distinctive cluster features. The partial aggregation strategy assumes that neural networks are divided into base layers and personalized layers to combat statistical heterogeneity \cite{arivazhagan2019federated}. When performing FL among cluster leaders, each leader uploads all or only a part of the trained weights to the server, while they receive only the updated weights for the base layers. 
    
    Let $B$ and $(L-B)$ be the number of base layers and personalized layers, respectively. Since base layers are typically the first ones in the neural network model, the weight update in the $(t+1)^{th}$ training round is given by:
\begin{equation}
\small
	\omega_{\rm gl}({t+1}) = \sum_{k=1}^{K}a_k \omega_{c_k}(t),
	\label{eq4}
\end{equation}
where $K$ is the number of cluster leaders participating in the FL round, $a_k \in [0,1]$ is the weight factor of cluster leader $c_k$ in the global aggregation such that $\sum_{k=1}^K a_k=1$, and $\omega_{\rm gl}$ (resp. $\omega_{c_k}$) represents the updated global (resp. local) neural network weights (resp. of leader $c_k$, $\forall k=1,\ldots,K$) of base layers. 
Once the global neural network model is updated, the FL server broadcasts the aggregation outcome for the $B$ base layers to the cluster leaders. The latter update their base layers weights as follows: 
\begin{equation}
\small
	\omega_{c_k}^{l}(t+1) = \omega_{\rm gl}^{l}(t+1),\quad \forall l \in [1,B],\; k=1,\ldots,K.
	\label{eq6}
\end{equation}
The above process is repeated for $T$ FL training rounds, as described in Algorithm \ref{algo1}, and where the function $\textbf{Louvain}$ is the clustering algorithm. 
\end{itemize}



\begin{algorithm}[h!]
\footnotesize
	\caption{Federated Learning Session}
	\label{algo1}
	\begin{algorithmic}[1]
		\renewcommand{\algorithmicrequire}{\textbf{Input:}}
		\renewcommand{\algorithmicensure}{\textbf{Output:}}
		\REQUIRE {Nbr. of clients $N$, nbr. of clusters $K$, nbr. of FL rounds $T$.}
		
		\BlankLine
		\textit{Initialization} :
		\STATE Get $\omega_i, \forall i=1,\ldots,N$ after a short local training
		
		\BlankLine
		\textit{Clients clustering}:
		\FOR{$i\leftarrow 1$ \KwTo $N-1$}
		\FOR{$j\leftarrow i+1$ \KwTo $N$}
		\STATE {Update similarity graph $G(V,E)$ using eq.(\ref{eq2})}
		\ENDFOR
		\ENDFOR
		\STATE {Get $\{c_1, \ldots, c_K\} \leftarrow$ \textbf{Louvain}$(G(V,E), K)$}
		
		\BlankLine
		\textit{Federated learning} :
		\WHILE{$t<T$}
		\STATE {Update the global neural network model based on eq.(\ref{eq4})}
		\STATE Broadcast $\omega_{\rm gl}(t+1)$ to cluster leaders
		\FOR{$k\leftarrow 1$ \KwTo $K$}
		\STATE{Update local model's base layers  using eq.(\ref{eq6})}
		\STATE Train local model for $\varepsilon$ episodes
		\ENDFOR
		\ENDWHILE		
		
		\BlankLine
		\ENSURE {Cluster leaders' neural network models.} 
		
	\end{algorithmic} 
\end{algorithm}


\vspace{-5pt}
\subsection{Transfer Learning Session}
After the $T$ rounds of FL, $c_k$'s neural network weights $\omega_{c_k}(T)$ include weights of base layers, trained through FL, and weights of personalized layers, trained only with the local data. 
Transfer learning consists on sending the pre-trained model weights of each leader to the members of its cluster. Consequently, members' models are initialized with the leader's model weights as follows:  
\begin{equation}
\small
	\omega_{j} = \omega_{c_k}(T),\quad \forall j\in C_k. 
	\label{eq5}
\end{equation}
Subsequently, each cluster model (other than the leader) starts training its model using its own dataset for at most $\eta$ episodes or until convergence.
This training process is equivalent to individual training and does not require any further communication among clients and/or FL server.



%
%
%
%
%

\vspace{-5pt}
\subsection{Communication Cost}
Communication cost of CEFL is decomposed into 4 parts:
\begin{enumerate}
	\item The upload of neural network weights of all clients at the short initial individual training to initialize clustering.
	\item The upload of base layers weights of leaders to the server in each FL round.
	\item The broadcast of base layers' weights from the server to leaders in each FL round.
	\item The transmission of all model weights from each leader to its cluster members in the transfer learning session.
\end{enumerate}

Let $\delta_l$ be the data size (in bits) of the weights in layer $l$ of the neural network model of any client/server. Hence, the total amount of data transiting in the FL system, denoted by $\Delta$, is given by
\small
\begin{eqnarray}
		\Delta &=& N \sum_{l=1}^{L}\delta_l+ K T \sum_{l=1}^{B}\delta_l+T \sum_{l=1}^{B}\delta_l+K \sum_{l=1}^{L}\delta_l\nonumber \\
		&=& 
		\left( N+K \right) \sum_{l=1}^L \delta_l+ 		T \left( K+1 \right) \sum_{l=1}^B \delta_l \; \rm{(bits)}.
		\label{eq7}
\end{eqnarray}
\normalsize
\noindent
Assuming that each bit has a unitary cost, then the communication cost is equal to $\Delta$.  

	%
	%
	%
	%
	%

\section{Experimental Evaluation}
\label{sec 4}


\subsection{Dataset and Preprocessing}
The FL experiments conducted on this work are related to a health monitoring application, where collected data from patients is analyzed to identify their activities. Specifically, we relied on the public dataset MobiAct\cite{vavoulas2016mobiact}. MobiAct is a dataset for activity recognition, where data from 67 patients, between the ages of 20 and 47, is obtained and labelled. Data is collected using the patients' smartphones when they are performing different activities. The application focuses on four types of fall activities and nine types of daily activities. For the sake of simplicity, we decide to focus only on activities that indicate possible upcoming falling, thus reducing the number of recognizable classes to eight types only, namely the initial fall activity classes, i.e., forward-lying, front-knees-lying, sideward-lying, and sack-sitting-chair, three fall-like activity classes, i.e., sit chair, car step in, car step out, and one daily activity class including all of standing, walking, jogging, jumping, stairs up and stairs down types.


We opt here for the data preprocessing method proposed in \cite{he2019low}, which samples 3-axial accelerations and angular velocities data using sliding windows and converts them into the bitmap format. {{Given a subject's 3-axial sampled signal of one activity, a sliding window with a given slide interval $I_{type}$ that moves along the entire signal is used to capture signal features\footnote{The slide interval refers to the number of sampling points between the starting points of two successive windows.}. Data captured by each sliding window is used to construct a red-green-blue (RGB) bitmap image, where data of accelerations and angular velocities from one axis are taken as pixel RGB values.}} For efficiency purposes, we optimize the slide interval size between windows.
{{Indeed, in the MobiAct dataset, different types of activities are recorded for different time durations, denoted $t_{type}$. Since all fall activities are sampled for 10 seconds, we empirically initialize the reference sliding interval $I_0$ to be 40. However, some activities, such as walking, are recorded for up to 10 minutes. In order to avoid making processed dataset more unbalanced, we propose to adjust the sliding intervals for different activities, called $I_{type}$, to their recorded duration. Let $t_0 = 10$ seconds be the reference duration, then, the slide interval for each activity should be customized according to 
\begin{equation}
\small
I_{\rm type} = I_0  \frac{t_{\rm type}}{t_0}.
\label{eq8}
\end{equation}
}} 

\subsection{Experimental Setup}
The neural network model used in this paper is the fall detection convolutional neural network (FD-CNN) proposed in \cite{he2019low}. FD-CNN takes as input the 3-channel $20\times20$ RGB bitmap image. It is composed of 2 convolutional layers, 2 subsampling layers, and 2 fully-connected layers. The filter size of the 2 convolutional layers is $5\times5$, while the filter numbers are 3 and 32, respectively. A $2\times2$ max-pooling layer follows each of the convolutional layers, while the fully connected layers include 512 and 8 units, respectively. FD-CNN adopts ReLU as the activation function, while the softmax function is used in the last fully-connected layer to normalize the output to a probability distribution. The learning rate is set to $10^{-4}$. The neural network is trained by the Adam optimizer with a batch size of 32, and the cross-entropy loss function is adopted to measure the classification performance. Finally, we set $a_k=1/K$, $\forall k=1,\ldots,K$.

For our experiments, we compare the performance of CEFL with the following baselines:
\begin{itemize}
\item \textbf{Regular FL:} It is the conventional federated learning between the server and all clients, where FD-CNN is the same training model for the server and clients.
\item \textbf{FedPer:} It is a federated learning approach with base layers and personalized layers that combats the degradation from statistical heterogeneity \cite{arivazhagan2019federated}. The neural network model for the server and clients is FD-CNN.
\item \textbf{Individual Training:} Training is conducted by the clients themselves without any data exchange or model communication. The neural network of each client is FD-CNN.
\end{itemize} 


\subsection{Results and Discussion}
\begin{figure}[t]
\centering
\includegraphics[width=0.4\textwidth]{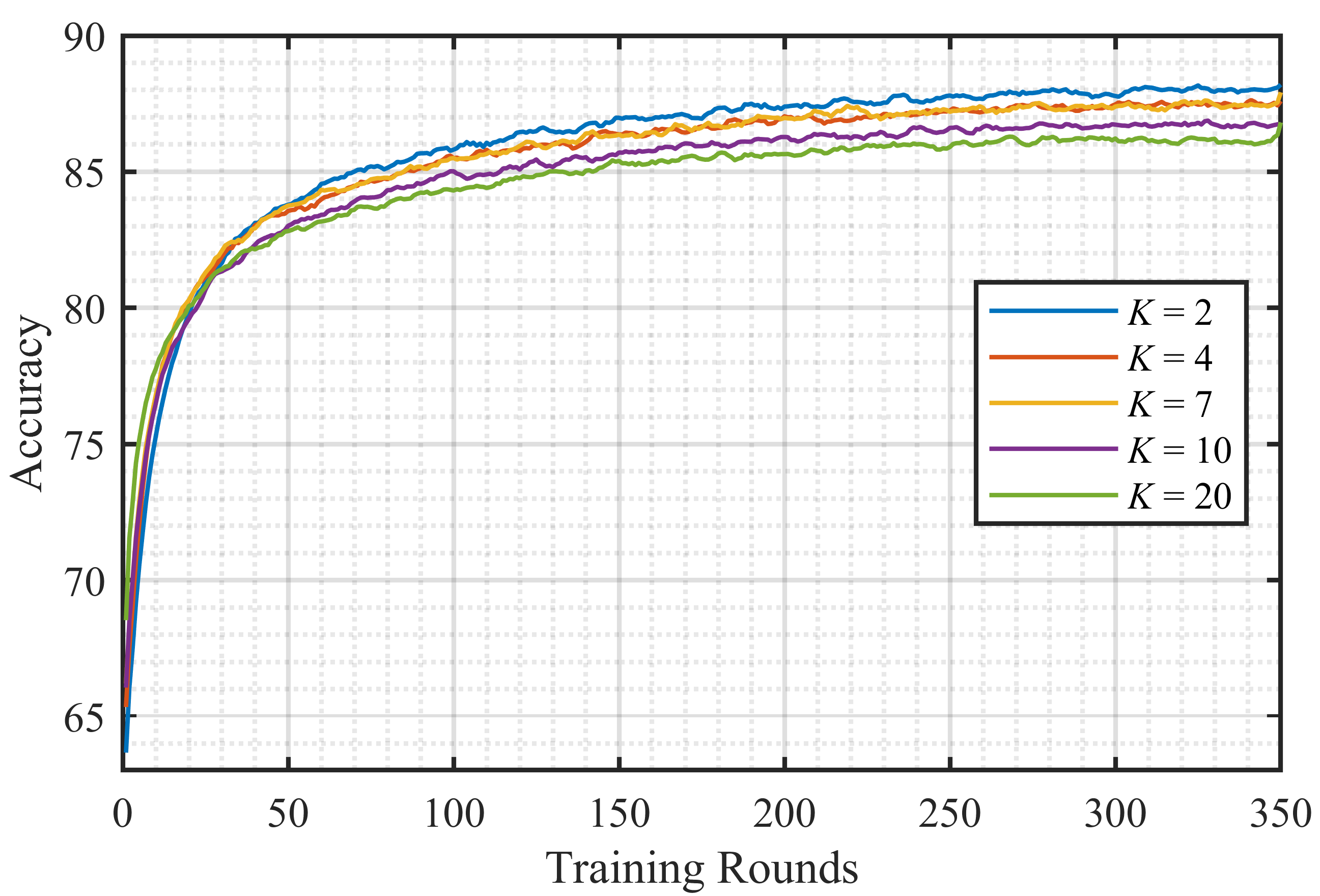}
\caption{CEFL accuracy vs. nbr. of rounds (different $K$).}
\label{fig3}    
\end{figure}

First, in the proposed CEFL framework, the number of clusters $K$ is an important parameter to determine. Its choice might influence how representative the chosen leaders are. To clarify its impact, we evaluate in Fig. \ref{fig3} the FL accuracy performance, calculated as the average clients' accuracy, for different $K$ values. When $K$ grows from 2 to 20, accuracy gradually reduces from 88.2\% to 86.81\%, making $K=2$ the optimal number of clusters in CEFL. Consequently, we fix $K=2$ for the remaining experiments.


In Table \ref{tab1}, we compare the proposed CEFL to the aforementioned baselines in terms of complexity (number of training/aggregation rounds/episodes), accuracy, and communication cost (in megabytes - MB).
Although Regular FL presents the best accuracy, its communication cost is the highest due to frequent model updates between the server and clients. Through partial-FL aggregation, FedPer reduces the communication cost by 0.5\% only compared to Regular FL, while the accuracy drops from 91.07\% to 88.78\%, which is not consistent with the performance illustrated in \cite{arivazhagan2019federated}. This outcome might result from the dataset type and applied changes in the preprocessing step. Both Regular FL and FedPer run for $350 \times 8=2800$ training episodes. In contrast, Individual training requires no communications, however, it reaches a low accuracy of 84.86\% due to the limited scale of local datasets. In terms of complexity, it runs the lowest number of episodes equal to 350. Finally, the proposed CEFL cuts the communication cost down from 79730 MB to 1231 MB when compared to Regular FL, which is a 98.45\% in cost savings. This significant reduction comes at the price of a slightly lower accuracy of 88.2\%, which is comparable to the FedPer performance, but still at a fraction of the communication cost. Also, CEFL runs for $100 \times 8 + 350 = 1150$ episodes that is 60\% lower than the one for Regular FL and FedPer. 

\begin{table*}[t]
\caption{Comparison of different training models}
\begin{center}
	\begin{tabular}{|p{2.6cm}|p{2cm}<{\centering}|p{2cm}<{\centering}|p{3.2cm}<{\centering}|p{2cm}<{\centering}|p{3.5cm}<{\centering}|}
		\hline
		\multirow{3}{*}{\textbf{Method}}&\multicolumn{3}{p{6cm}<{\centering}|}{\textbf{Training Rounds}}& \multirow{3}{*}{\textbf{Accuracy (\%)}}& \multirow{3}{*}{\makecell{\textbf{Communication}\\ \textbf{Cost} (MB)} }\\
		\cline{2-4} 
		&\multicolumn{2}{p{4cm}<{\centering}|}{\textbf{Federated Learning}}& \multirow{2}{*}{ \makecell{\textbf{Local}\\ \textbf{episodes}$^{\mathrm{b}}$} }& & \\ \cline{2-3}
		& \makecell{Aggregation\\ rounds} & \makecell{Local episodes\\ per aggregation} &&&\\ \hline
		Regular FL  & 350 & 8 &--&91.07& 79730 \\ \hline
		FedPer  & 350 & 8 &--&88.78& 79357\\ \hline
		Individual Training  & -- $^{\mathrm{a}}$ & -- &350&84.86&0 \\ \hline
		\textbf{CEFL}  & \textbf{100} & \textbf{8} &\textbf{350}&\textbf{88.20}& \textbf{1231}\\ \hline
		\multicolumn{4}{l}{$^{\mathrm{a}}$ \textbf{--} means not applicable. {$^{\mathrm{b}}$ Local training occurs outside from the FL process.}} \\
	\end{tabular}
	\label{tab1}
\end{center}
\end{table*}

\begin{figure}[t]
\centering
\includegraphics[width=0.4\textwidth]{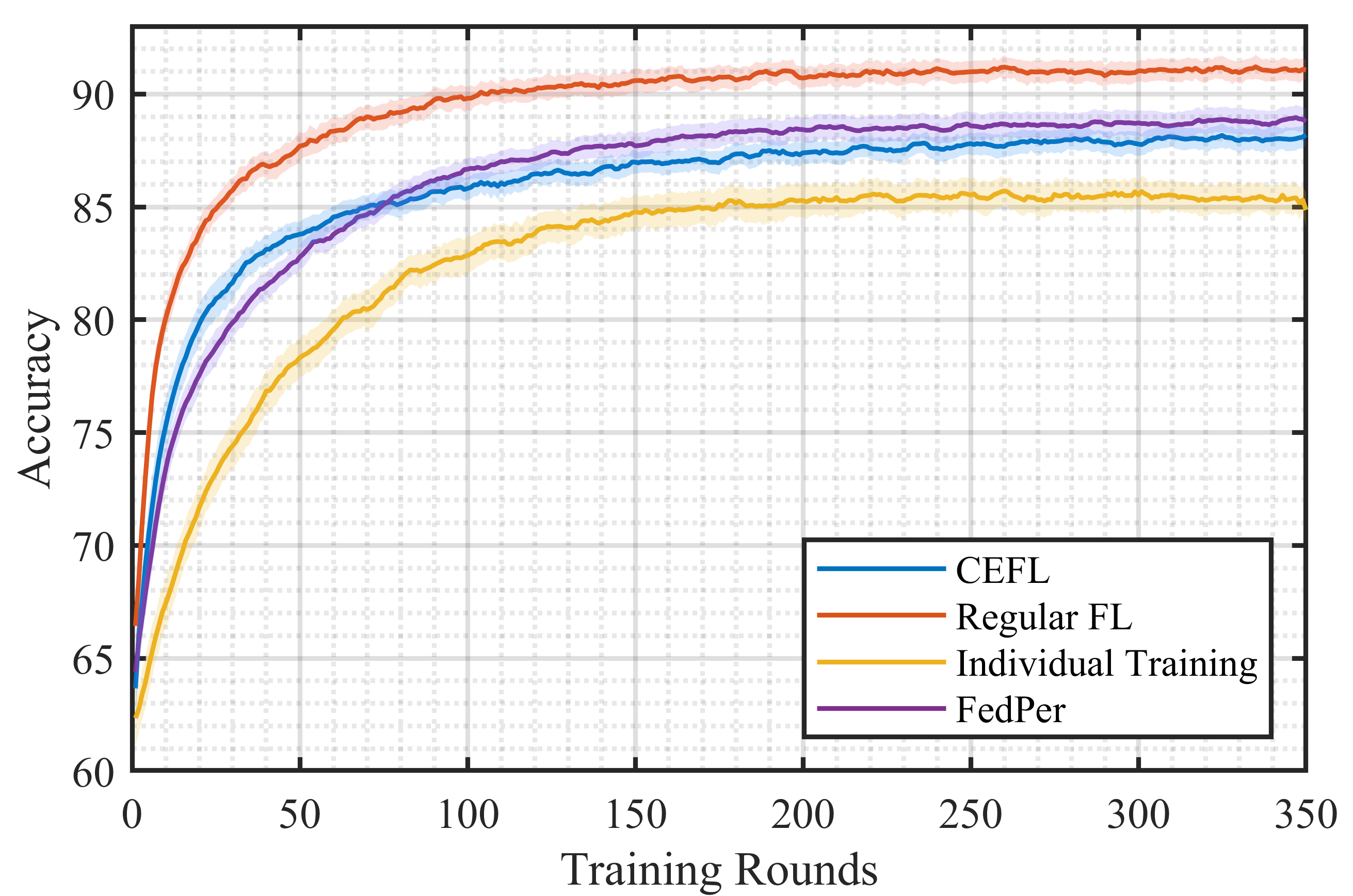}
\caption{Accuracy vs. nbr. of rounds (different training methods). 
}
\label{fig4}    
\end{figure}

The convergence behaviour of the aforementioned methods is depicted in Fig. \ref{fig4}. Regular FL converges the fastest due to the continuous participation of all clients in the training process. Our method CEFL converges also fast due to transfer learning between leaders and cluster members. In contrast, FedPer converges slowly since there is no information sharing for the personalized layers. Finally, Individual Training converges the slowest due to the independent operation of clients. The shaded areas around curves indicate the standard deviation of accuracy. As it can be seen, CEFL and Regular FL demonstrate the most stable performance behaviour, while the remaining two methods have a higher deviation, which indicates an unstable convergence trend.

\begin{figure*} 
\centering
\subfloat[Client 4]{
	\includegraphics[width=0.32\linewidth]{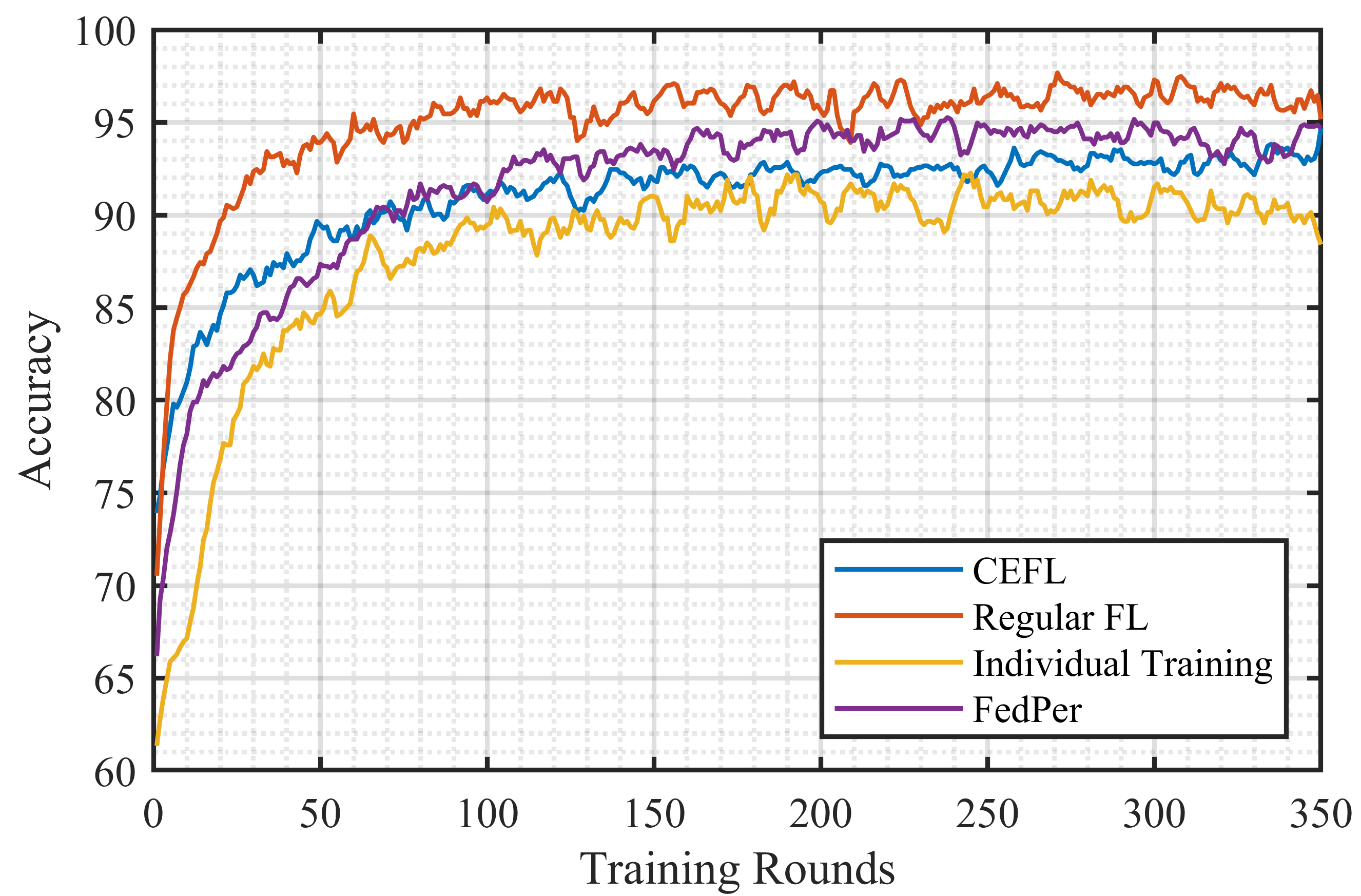}}
\label{5a}\hfill
\subfloat[Client 31]{
	\includegraphics[width=0.32\linewidth]{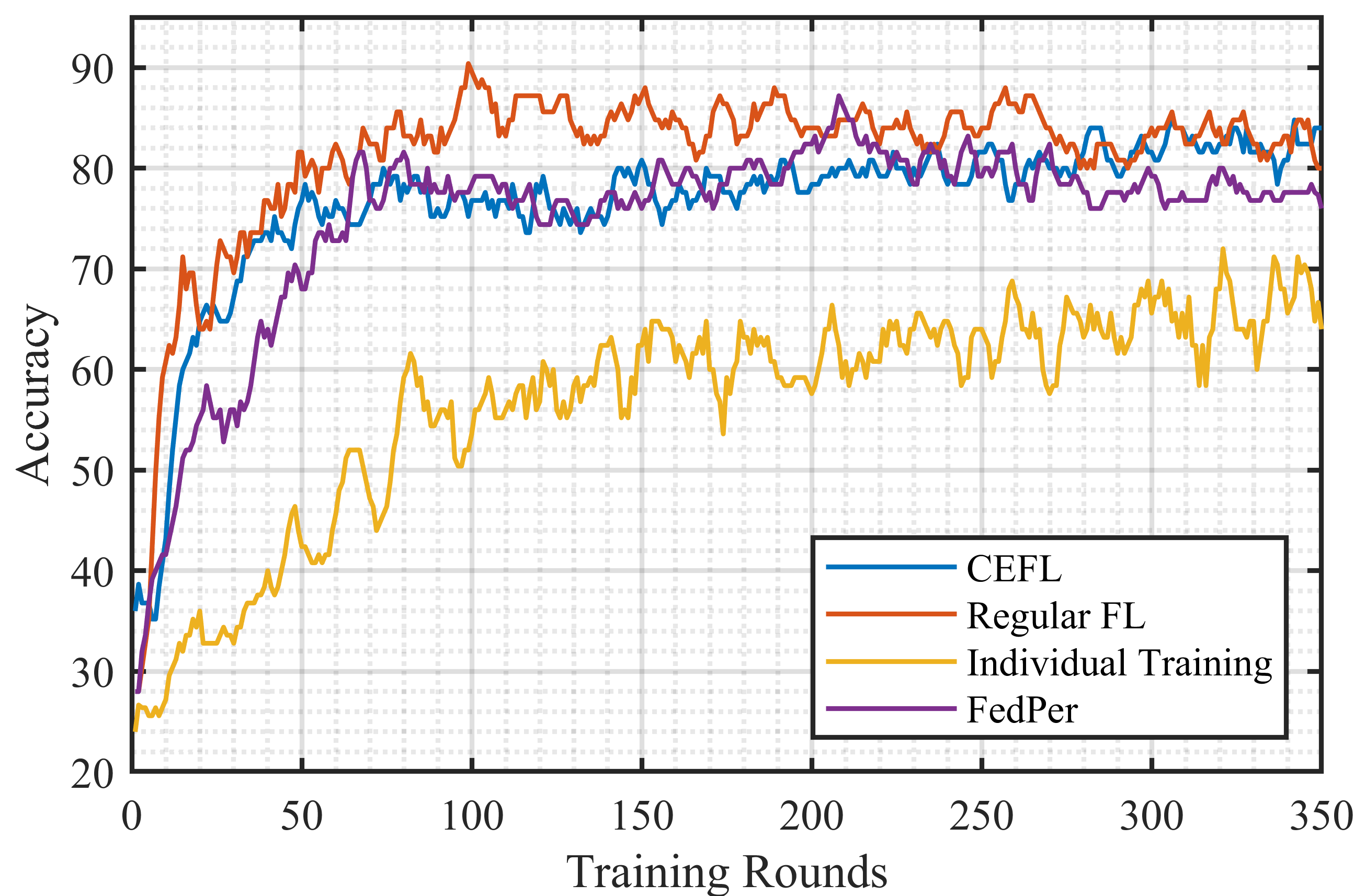}}
\label{5b}
\subfloat[Client 50]{
	\includegraphics[width=0.32\linewidth]{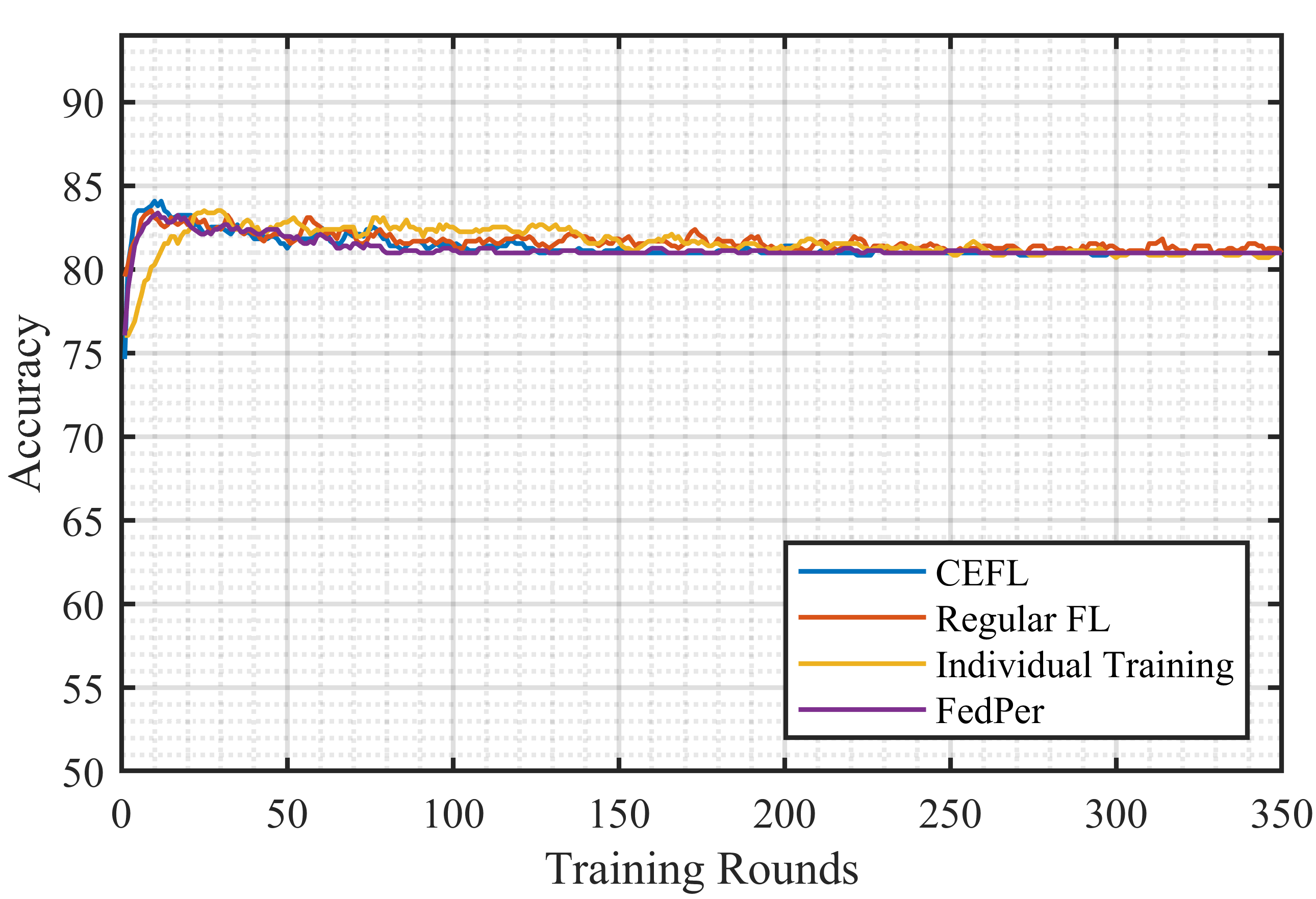}}
\label{5c}
\caption{Accuracy convergence for three clients with different dataset distributions.}
\label{fig5} 
\end{figure*}

To study the impact of dataset heterogeneity, we evaluate the accuracy performances of 3 clients with different dataset features in Fig. \ref{fig5}. Specifically, we select clients 4, 31, and 50 characterized as follows: Client 4 owns 831 training samples that are representative of all 8 activity classes; Client 31 has only 101 training samples that are from the four types of fall activities only; and Client 50 has 570 training samples, with a predominance of 431 samples from a single activity class (daily activity). From Fig. \ref{fig5}, we notice that the accuracy of Client 4 is the highest compared to others. This is expected since it has the highest number and most representative data distribution among the targeted classes. In contrast, the accuracy of Client 31 is the worst due to its small-sized and unbalanced dataset. Although Client 50 has a relatively large-sized dataset, its unbalance significantly impacts the accuracy performance. Nevertheless, the accuracy of the different methods are almost the same for Client 50, while a significant gap is present for Client 31.     
Consequently, we conclude that our method has a similar performance to Regular FL when the client's dataset is small or highly unbalanced, while for clients with large or relatively balanced datasets, a slight performance gap can be noticed as in Fig. \ref{fig3}.


\vspace{-5pt}
\section{Conclusion}
\label{sec 5}
This paper proposes CEFL for health monitoring. CEFL consists of two steps: FL among cluster leaders and transfer learning from leaders to cluster members. Our method CEFL can reduce communication costs due to its small-scale FL among only selected leaders, selected through graph clustering and based on clients' mutual similarity. By inheriting trained models from the corresponding leader, cluster members can rapidly achieve an acceptable accuracy on their own datasets. Compared to baselines, CEFL achieves a great balance between communication and accuracy. 
Specifically, the communication cost can be reduced up to 98.45\% at the expense of less than 3\% accuracy degradation, compared to the best baseline. Moreover, for clients with small or highly imbalanced datasets, CEFL yields as high accuracy as the best baseline with at a fraction of the communication cost. 

\vspace{-5pt}
\section*{Acknowledgment}
\small
This research is supported by the Natural Sciences and Engineering Research Council of Canada, Huawei Canada, and a MITACS Globalink scholarship.

\vspace{-5pt}
\bibliography{IEEEabrv,citation.bib}

\begin{thebibliography}{10}
\providecommand{\url}[1]{#1}
\csname url@samestyle\endcsname
\providecommand{\newblock}{\relax}
\providecommand{\bibinfo}[2]{#2}
\providecommand{\BIBentrySTDinterwordspacing}{\spaceskip=0pt\relax}
\providecommand{\BIBentryALTinterwordstretchfactor}{4}
\providecommand{\BIBentryALTinterwordspacing}{\spaceskip=\fontdimen2\font plus
\BIBentryALTinterwordstretchfactor\fontdimen3\font minus
  \fontdimen4\font\relax}
\providecommand{\BIBforeignlanguage}[2]{{%
\expandafter\ifx\csname l@#1\endcsname\relax
\typeout{** WARNING: IEEEtran.bst: No hyphenation pattern has been}%
\typeout{** loaded for the language `#1'. Using the pattern for}%
\typeout{** the default language instead.}%
\else
\language=\csname l@#1\endcsname
\fi
#2}}
\providecommand{\BIBdecl}{\relax}
\BIBdecl

\bibitem{selvaraj2020challenges}
S.~Selvaraj and S.~Sundaravaradhan, ``Challenges and opportunities in {IoT}
  healthcare systems: a systematic review,'' \emph{SN Appl. Sci.}, vol.~2,
  no.~1, pp. 1--8, Jan. 2020.

\bibitem{rahman2020defending}
M.~S. Rahman, N.~C. Peeri, N.~Shrestha, R.~Zaki, U.~Haque, and S.~H. Ab~Hamid,
  ``Defending against the novel coronavirus {(COVID-19)} outbreak: How can the
  internet of things {(IoT)} help to save the world?'' \emph{Health Policy
  Technol.}, vol.~9, no.~2, p. 136, Jun. 2020.

\bibitem{verma2018cloud}
P.~Verma, S.~K. Sood, and S.~Kalra, ``Cloud-centric {IoT} based student
  healthcare monitoring framework,'' \emph{J. Ambient Intelli. Humanized
  Comput.}, vol.~9, no.~5, pp. 1293--1309, Jan. 2018.

\bibitem{li2020federated}
T.~Li, A.~K. Sahu, A.~Talwalkar, and V.~Smith, ``Federated learning:
  Challenges, methods, and future directions,'' \emph{IEEE Sig. Process. Mag.},
  vol.~37, no.~3, pp. 50--60, May 2020.

\bibitem{liu2018fadl}
D.~Liu, T.~Miller, R.~Sayeed, and K.~D. Mandl, ``{FADL}: Federated-autonomous
  deep learning for distributed electronic health record,'' \emph{arXiv
  preprint arXiv:1811.11400}, 2018.

\bibitem{duan2019astraea}
M.~Duan, D.~Liu, X.~Chen, Y.~Tan, J.~Ren, L.~Qiao, and L.~Liang, ``Astraea:
  Self-balancing federated learning for improving classification accuracy of
  mobile deep learning applications,'' in \emph{Proc. IEEE Int. Conf. Comp.
  Design (ICCD)}, 2019, pp. 246--254.

\bibitem{Wu2020}
Q.~Wu, X.~Chen, Z.~Zhou, and J.~Zhang, ``{FedHome}: Cloud-edge based
  personalized federated learning for in-home health monitoring,'' \emph{IEEE
  Trans. Mob. Comput. (Early Access)}, pp. 1--1, 2020.

\bibitem{sozinov2018human}
K.~Sozinov, V.~Vlassov, and S.~Girdzijauskas, ``Human activity recognition
  using federated learning,'' in \emph{Proc. IEEE Int. Conf. Parallel \& Dist.
  Process. Apps. (ISPA)}, 2018, pp. 1103--1111.

\bibitem{fang2020highly}
C.~Fang, Y.~Guo, N.~Wang, and A.~Ju, ``Highly efficient federated learning with
  strong privacy preservation in cloud computing,'' \emph{Computers \&
  Security}, vol.~96, p. 101889, Sep. 2020.

\bibitem{konevcny2016federated}
J.~Kone{\v{c}}n{\`y}, H.~B. McMahan, F.~X. Yu, P.~Richt{\'a}rik, A.~T. Suresh,
  and D.~Bacon, ``Federated learning: Strategies for improving communication
  efficiency,'' \emph{arXiv preprint arXiv:1610.05492}, 2016.

\bibitem{sattler2019robust}
F.~Sattler, S.~Wiedemann, K.-R. M{\"u}ller, and W.~Samek, ``Robust and
  communication-efficient federated learning from non-{IID} data,'' \emph{IEEE
  Trans. Neural Nets. Learn. Syst.}, vol.~31, no.~9, pp. 3400--3413, Sep. 2019.

\bibitem{Amirho2020}
R.~Amirhossein, M.~Aryan, H.~Hamed, J.~Ali, and P.~Ramtin, ``{FedPAQ}: A
  communication-efficient federated learning method with periodic averaging and
  quantization,'' \emph{arXiv preprint arXiv:1909.13014}, 2020.

\bibitem{luping2019cmfl}
W.~Luping, W.~Wei, and L.~Bo, ``Cmfl: Mitigating communication overhead for
  federated learning,'' in \emph{Proc. IEEE Int. Conf. Dist. Comput. Syst.
  (ICDCS)}, 2019, pp. 954--964.

\bibitem{blondel2008fast}
V.~D. Blondel, J.-L. Guillaume, R.~Lambiotte, and E.~Lefebvre, ``Fast unfolding
  of communities in large networks,'' \emph{J. Stat. Mechanics: Theory and
  Experiment}, vol. 2008, no.~10, p. P10008, Oct. 2008.

\bibitem{arivazhagan2019federated}
M.~G. Arivazhagan, V.~Aggarwal, A.~K. Singh, and S.~Choudhary, ``Federated
  learning with personalization layers,'' \emph{arXiv preprint
  arXiv:1912.00818}, 2019.

\bibitem{vavoulas2016mobiact}
G.~Vavoulas, C.~Chatzaki, T.~Malliotakis, M.~Pediaditis, and M.~Tsiknakis,
  ``The mobiact dataset: Recognition of activities of daily living using
  smartphones,'' in \emph{Proc. Int. Conf. Info. Commun. Technol. Ageing Well
  and e-Health}, vol.~2, 2016, pp. 143--151.

\bibitem{he2019low}
J.~He, Z.~Zhang, X.~Wang, and S.~Yang, ``A low power fall sensing technology
  based on {FD-CNN},'' \emph{IEEE Sensors J.}, vol.~19, no.~13, pp. 5110--5118,
  Jul. 2019.

\end{thebibliography}
\bibliographystyle{IEEEtran}

\end{document}